# Predicate learning in neural systems: Discovering latent generative structures

Andrea E. Martin[1*] [orcid: 0000-0002-3395-7234]

Leonidas A. A. Doumas[2]

[1]Max Planck Institute for Psycholinguistics
[2]University of Edinburgh

1 October 2018

DRAFT: DO NOT QUOTE WITHOUT PERMISSION

*Corresponding author, contact details:
Andrea E. Martin
Max Planck Institute for Psycholinguistics
Wundtlaan 1
6525XD Nijmegen
The Netherlands
andrea.martin@mpi.nl
+31 (0) 243 521 585

ABSTRACT

Humans learn complex latent structures from their environments (e.g., natural language, mathematics, music, social hierarchies). In cognitive science and cognitive neuroscience, models that infer higher-order structures from sensory or first-order representations have been proposed to account for the complexity and flexibility of human behavior. But how do the structures that these models invoke arise in neural systems in the first place? To answer this question, we explain how a system can learn latent representational structures (i.e., predicates) from experience with wholly unstructured data. During the process of *predicate learning*, an artificial neural network exploits the naturally occurring dynamic properties of distributed computing across neuronal assemblies in order to learn predicates, but also to combine them compositionally, two computational aspects which appear to be necessary for human behavior as per formal theories in multiple domains. We describe how predicates can be combined generatively using neural oscillations to achieve human-like extrapolation and compositionality in an artificial neural network. The ability to learn predicates from experience, to represent structures compositionally, and to extrapolate to unseen data offers an inroads to understanding and modeling the most complex human behaviors.

## INTRODUCTION

As humans, we recognize our home, pet, or partner regardless of our viewing angle and the concomitant variation in the 2-D image on our retinas (e.g., [1]). Similarly, when we listen to speech or view sign, we understand linguistic structures that go far beyond any physical description of the stimulus (e.g., [2-4]). Furthermore, we have the capacity to promiscuously apply what we know to new situations, for example, if we have to improvise a recipe with novel ingredients, we would never entertain cooking something by refrigerating it.[1] These examples emphasize two things. First, the ability to use 'incomplete' or partial sensory experience to infer latent structures in the environment, and then reason and generalize based on them, appears to be crucial for everyday human behavior. Second, the domains where humans wildly outperform artificial intelligence systems (AI) seem to involve inference beyond lower order statistical relationships [5]. While it is clear that, in the limit, AI can outmatch human performance on pure computation and statistical tasks (e.g., medical imaging), it is not clear how domains that require inference (e.g., analogy, scene comprehension), decision making (e.g., diagnosis, game play), or abstract rule generation (e.g., natural language) can be approached without a profound change in the principles of computation currently being espoused in both cognitive science and AI (for a discussion see [4-7].

In this opinion paper, we argue that *the ability to learn representational structure from experience* and, crucially, *to represent structures independently from the particular data and context in which they were learned* underlies a system's capacity to generalize its knowledge to new experiences, contexts, and untrained stimuli in an extrapolatory, human-like way [8,9]. We summarize the computational principles needed to achieve structure (i.e., compositional or predicate representations) and variable-value independence [5] in a distributed computing system in an approach we call *predicate learning* (Figure 1). Predicate learning represents the integration of formal symbolic models with traditional neural computing principles and capitalizes on the information carried by oscillatory rhythms of neuronal computation.

[1] For recipes generated by deep learning networks, seehttps://www.dailydot.com/unclick/neural-network-recipe-generator/

Advances in AI and machine learning [10] have produced deep neural network (DNN) systems that reach and even exceed human levels of performance on a range of cognitive tasks [11]. DNNs can learn to perform a variety of tasks without any prior representations or knowledge (e.g., to play an Atari video game from pixel data), but it is well known that they struggle with tasks that require generalization to input from outside the bounds of the training set (c.f., ranging from object recognition, inference, analogy, language; [5]). It is likely that DNNs' explicit lack of structured representations plays a role in this struggle because accounts of how humans generalize tend to rely on powerful symbolic languages formed from structured representations (predicates), which can be generatively applied to new arguments [9,12-15]. Structured representations like predicates allow the flexible transfer of information across contexts because the same representations can be used to effectively characterize wildly different input data (e.g., the relational predicate "beside" can be applied to tennis balls in play over nets and to roofs on houses). But current models that exploit structured representations face a complementary challenge compared to DNNs: They require specification, by the modeler, of a collection of necessary representational structures in advance of any actual learning; in other words, they do not learn the contents of their structures directly from the environment without the use of pre-specified representations and rules [cf. 9,13,14,15]. That is, while structure-based models generalize more flexibly than DNNs, they do not perform general "from scratch" learning and often feature powerful symbolic languages that were hand-coded by the modeller [e.g., 9,14,15]. As a result, structured models often make strong nativist claims, for example, that a large set of representational elements and the rules for building compositions of these elements must be innate [16].

Providing an account of how structured representations like predicates can be learned from experience has proven difficult (e.g., [17]). But, in our view, a demonstration that predicates can be learned from experience, and ergo, that information learned in one context can be generalized to untrained inputs and contexts, would have substantial impact on theories of human reasoning, language representation and processing, and on applications in AI. In order to explain how structure is learned, we give a minimum definition of what structure is. If structure learned from the environment is crucial to human cognition, why is it so hard for a machine system to learn?

***Three requirements for structured representations***

It is important to be clear about what we mean by *representations* and *representational systems*. A *representation* is a state of a system that explicitly carries information about something else in the world, namely, about how the *representational elements* of that system carry information about that thing (e.g., a painting of a sandwich is a representation of that sandwich where the representational elements are paint and the configuration of the paint such that it resembles the sandwich; an equation is a representation of a particular mathematical system composed of representational elements like variables and operators in that system). Thus, a *representational system* minimally consists of two things: i) representational elements (e.g., nodes, weights, or connections in a neural network, symbols in a symbolic model, firing patterns, neurotransmitter dynamics, gene expression, network dynamics in biological neural systems) and ii) a set of rules or processes for inferring new states from existing states (see [18]). A *structured representational system* must further support the composition (e.g., binding) of basic representational elements into more complex structures in a manner that is dynamic (bindings must be created and destroyed on the fly) and that maintains the representational independence of the items so bound (e.g., [19,20,21]).

To summarize, for a system to be structured, it must include: (i) **states that carry information about the world via its representational elements** (i.e., specifying what is present in a given situation or dataset), (ii) **a mechanism that carries dynamic binding information** (specifying how those elements are arranged), and (iii) **processes by which new representational structures are inferred from existing structures**. Crucially, these three informational degrees of freedom must be independent (i.e., the binding operator must not affect the meaning of the bound items; [6,7]). Any system that can represent these three independent sources of information can, in principle, support structured representations. In the following example, we use distribution over the neural network space, activation values, and time as our three independent informational degrees of freedom.

***Predicate learning: detection of invariance through intersective comparison***

A system that can learn structure from experience bridges the gap between the general flexibility of deep learning and the formal power of structured systems. In this section, we briefly describe how a system can discover functionally symbolic representations from experience by exploiting basic neurophysiological mechanisms in a process we call *predicate learning*. The representations learned in this way allow a system to generalize and transfer that expertise in a human-like fashion, quickly (in one shot) and without explicit feedback [4,8,22] because predicate structures can be applied generatively and compositionally.

The architecture, called DORA (*Discovery of Relations by Analogy*; [4,8,22]), is descended from the symbolic-connectionist system LISA [13, 23] and is based on two fundamental concepts from cognitive science and neuroscience. The first concept is that learning and generalization depend upon a process of comparison [24]. The second is that information in neural computing systems can be carried by the oscillatory regularities that emerge as its component units fire [25,26]. Unsupervised comparison is used to discover which characteristics of the input are invariant (see Figure 2). The common pattern of distributed activation that occurs across many inputs to the network is discovered by calculating the intersection across the inputs' activation states. The distributed pattern that is common (i.e., the intersection itself) is a latent structure. The intersection becomes an activation state of the network that is separable from the inputs from which it was learned, and thus forms a functional predicate (when combined with dynamic binding) that reflects the latent invariants in the dataset or environment.

During predicate learning, the system uses oscillatory regularities in the network to dynamically bind predicates and arguments, without fundamentally changing the meaning of any elements so bound (though statistics about the binding can be tracked). Crucial to the comparison learning algorithm is the fact that the DORA architecture features banks of layered units that are connected to a common pool of feature units. Activation flows from one bank (e.g., the current attentional state of the model) to other banks (e.g., active memory and long-term memory) via these shared feature units. DORA maps (i.e., learns connections between) corresponding coactive units across banks using a modified Hebbian learning algorithm [13,23], allowing for comparison of activation states related to the input. In sum, the system assumes (a) a layered architecture, with banks of layered units connected to a common pool of feature units, (b) lateral inhibition between units in the layered banks, (c) yoked inhibitors on units that accumulate input from their yoked units, and units at higher layers, and (d) the capacity for Hebbian learning. There is ample evidence for these basic architectural and computational features in the human neuroscience literature [27]. As a result of these architectural principles, DORA learns representations that are functionally and formally symbolic from training data, without feedback, and without requiring that structured representations be specified a priori. Full details of the model’s operation appear in [8] and [22]. The model has been used to capture a wide range of phenomena in the development of relational reasoning and analogy [22], in the cortical tracking of linguistic structures [4], and to extrapolate predicates learned from one video game in order to play a different, untrained game [8].

***Neural oscillations as the rhythms of computation***

Predicate learning exploits a core set of neurophysiological computing principles, namely that computation in a neural network is rhythmic. Most crucially, predicates, once learned, are dynamically bound to their arguments by phase-lag, which is expressed as systematic asynchrony of unit firing [25,26,28], or desynchronization between the activation cycles of the nodes coding predicates and arguments (Figure 3). During asynchrony-based—or phase-lag-1—binding, as a predicate or proposition becomes active, bound arguments and predicates fire in direct sequence, and out of synchrony with other bound predicate-arguments sets. This feature is what allows the system to maintain independence between a predicate and its argument(s), and achieve variable-value independence [29]. At the same time, binding information is carried in the proximity of firing (e.g., with predicates firing directly before their arguments), meaning that representing predicates in a neural system relies critically on sensitivity to time, and rhythm, as dimensions of computation. Synchrony-based—or phase-lag-0—binding also occurs in the system depending on the computational goal, for example, a proposition can be activated by having its bound arguments and predicate-argument roles fire together, but out of synchrony with other bound role-filler sets, in order to perform propositional-level computation of higher arities. By grouping representations into phase sets, or what is in and out of phase in the network, the system uses the rhythms of computation to both separate and combine information as needed.

Cortical oscillations have long been implicated as the indices of neural information processing [30]. Predicate learning in an artificial neural network relies on exploiting the naturally occurring "neural" oscillations of distributed computation over time. Being sensitive to how information is carried in time in a neural system implies that the dynamics of the system can themselves be learned from. A similar principle appears in the dynamic reorganization of cortical networks during learning in humans (e.g., [31]). Using oscillatory assembly activation to compute and to learn is potentially transformative, not only for its computational power (e.g., being able to learn from past states and learn relations over multiple time points and states), but also for the mechanistic link to neuroscientific theory and data (neural oscillations), and to formal accounts of cognition, including formalism of natural language and predicate calculi [15,16,32].Computing with neural oscillations represents a fundamental formal and neurophysiological synthesis between how human-like representations can be achieved in an artificial system that learns, and how distributed neural computing systems, including cortical assemblies in biological brains, process information.

***Towards compositionality and generative representational systems***

Predicate learning offers an account of how complex concepts might develop in neural computation systems without the need to hardwire or encode a priori structure, a theoretical and implementational limitation of current structure-based accounts of cognition (e.g., [9,15,16]), and offers a solution to the classic generalization problem that unstructured deep-learning systems face (e.g., [5,11]). These advances also relate to both the formal and functional senses of *generative models* - formally, because the information in predicate representations can entail a joint probability distribution that can be used to determine a given variable's probability based on another's, and functionally, because learned predicate calculi supports both the inference that allows the user to postulate a probable cause for an observed signal and the capacity to generate representational structures, through the composition of existing representations, as per the demands of the environment and of human behavior. Generative models, whether structured or not, have long been important in cognitive science (e.g., [33,34]) and AI (e.g., [35]), and in neuroscience (e.g., [36]). However, similar to the extrapolatory generalization problem faced by DNNs, unstructured deep generative models require extensive labelling of training data [37] and superhuman training routines [5]. In contrast, a system that uses predicate learning can discover and predicate what is latent in the environment, and discover what is relevant for behavior.  Predicate learning ultimately relies on the capacity of a system to be compositional - to host representations that can be combined in order to generate new representations.

In sum, we have described in brief how predicates can be learned from unstructured data using intersective comparison and rhythmic, desynchronized neural oscillations. Learning symbolic structure from signals that naturally occur in distributed computing systems offers a promising approach whereby the computational principles that can yield the highest forms of the human mind (e.g., relational reasoning, formal and natural language processing) can also be realized in systems based on the computational primitives of neurophysiology.

## Figure Captions

**Figure 1. Visual summary of the main concepts, claims, and definitions in the text**

**Figure 2. Cartoon illustration of predicate learning through intersection discovery**

**Figure 3. Cartoon illustration of rhythmic computation of predicate-argument structures through neuronal assembly desynchronization**

neurally implementable → formally structured

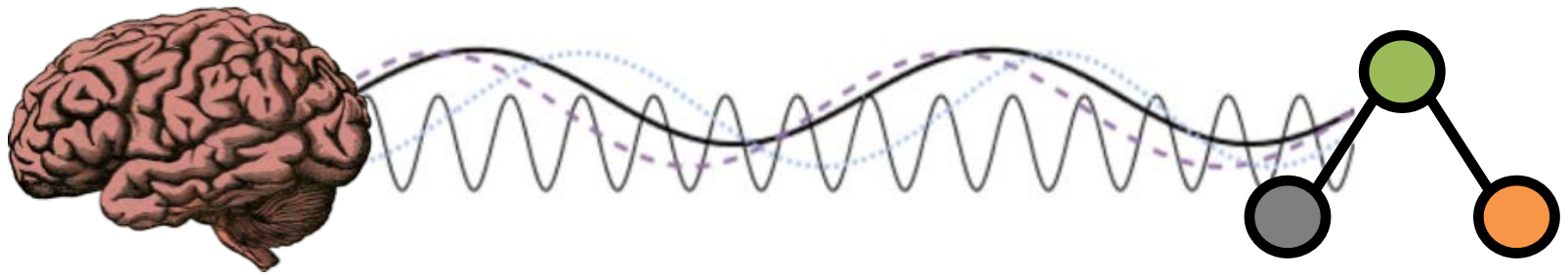

Neural Computing
(learns from data)

Predicate Learning
(learns to represent symbolic languages)

Bayesian Models
(represent symbolic languages)

**The core claims of predicate learning**

- learning structured representations that support **extrapolatory generalization** and **compositionality** in neural systems are crucial for human behavior
- functionally symbolic structures can be learned from unstructured data using **sensitivity to time** as a informational degree of freedom
- predicate learning can be implemented via **desynchronization** of neuronal assemblies

**predicate learning:** a series of algorithms that perform intersective comparison and uses rhythmic computation to discover latent structures

**rhythmic computation:** the exploitation of endogenous neural oscillations, such as (de)synchronization of aggregate unit firing over time, to compute separable information types over multiple timescales.

Figure 1.

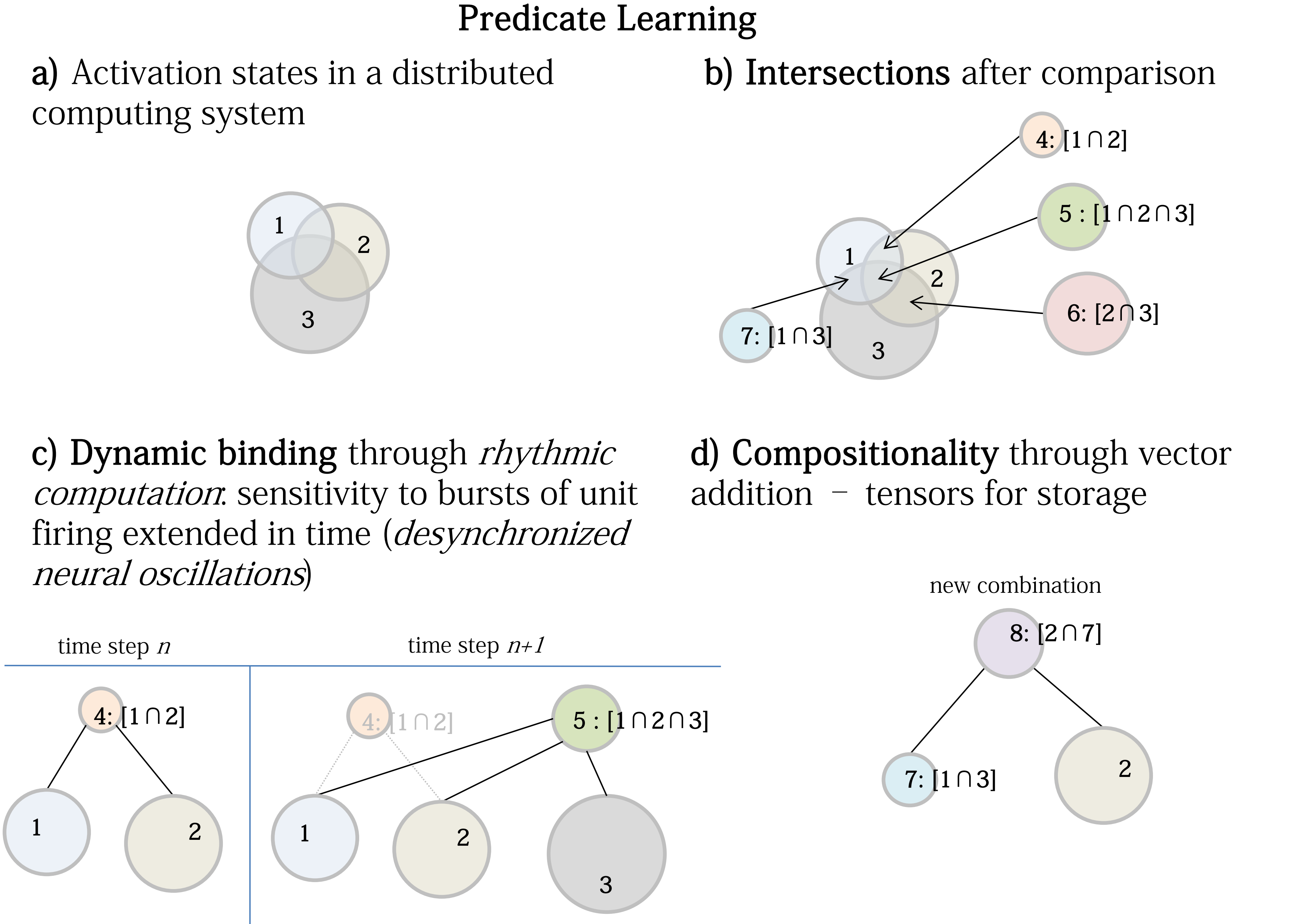
Predicate Learning
a) Activation states in a distributed computing system
1
2
3
b) Intersections after comparison
4: [1∩2]
5 : [1∩2∩3]
6: [2∩3]
7: [1∩3]
1
2
3
c) Dynamic binding through rhythmic computation: sensitivity to bursts of unit firing extended in time (desynchronized neural oscillations)
time step n
time step n+1
4: [1∩2]
1
2
4: [1∩2]
5 : [1∩2∩3]
1
2
3
d) Compositionality through vector addition – tensors for storage
new combination
8: [2∩7]
7: [1∩3]
2

Figure 2.

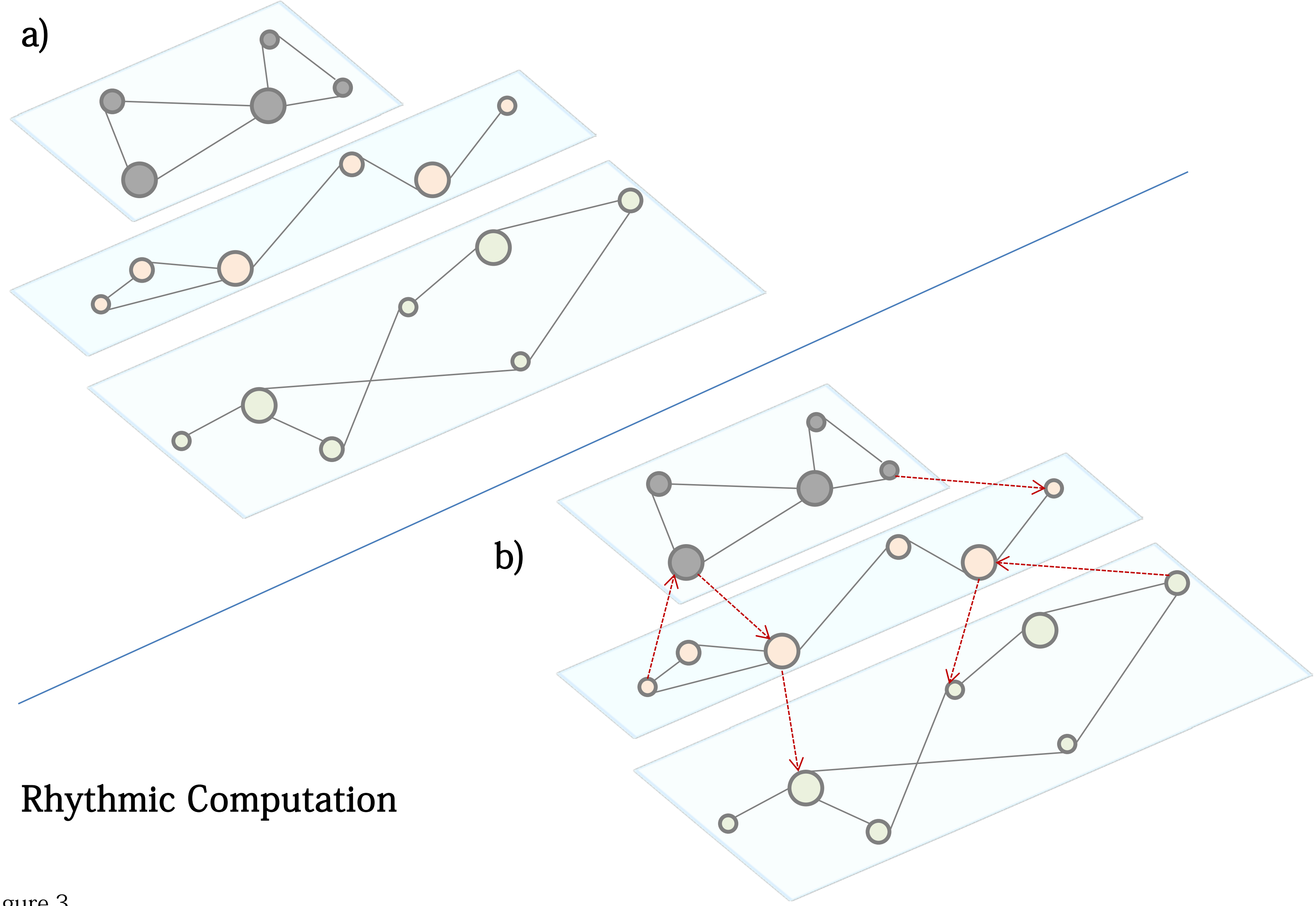

Figure 3.